\documentclass[runningheads]{svproc}

\usepackage{amsfonts}
\usepackage{latexsym}
\usepackage{amssymb}
\usepackage{amsmath}
\usepackage[mathscr]{eucal}

\renewcommand{\qed}{\hfill{\ \ \rule{2mm}{2mm}} \vspace{0.2in}}

\newcommand{\ind}{1\hspace{-2.3mm}{1}}

\renewcommand{\thefigure}{\arabic{figure}}
\begin{document}

\title{Ensemble Bounds for Pairwise Dissimilar Decomposition of Deterministic Datasets}
\titlerunning{Ensemble Decomposition Bounds for Datasets}


\author{ \textbf{Ghurumuruhan Ganesan}}
\authorrunning{G. Ganesan}
\institute{University of Bristol, UK\\
\email{gganesan82@gmail.com }}


\date{}
\maketitle

\begin{abstract}
In this paper, we study reduced similarity decompositions of \emph{deterministic} datasets using redundancy graphs and obtain \emph{ensemble} bounds for the minimum size of such constrained decomposition. We consider batch decompositions that reduce similarity \emph{across} batches and use the Lov\'asz local lemma to obtain bounds for the minimum size of a decomposition with a given pairwise dissimilarity. We then consider unique batch decomposition with a given relaxation parameter and obtain upper and lower ensemble bounds for the minimum size. Throughout, we use Binomial random graphs to obtain the desired ensemble bounds, by conditioning on the number of similar \emph{pairs} of data points.

\vspace{0.1in} \noindent \textbf{Key words:} Pairwise Dissimilar Decompositions; Unique Batch Decomposition; Deterministic datasets; Reduced Similarity; Ensemble bounds;

\vspace{0.1in} \noindent \textbf{AMS 2000 Subject Classification:} Primary: 60K35, 60J10;
\end{abstract}

\bigskip

\renewcommand{\theequation}{\thesection.\arabic{equation}}
\setcounter{equation}{0}
\section{Introduction} \label{intro}
Large datasets are typically broken down into small batches before being fed to a predictive model like neural network, in order to improve the learning performance and also ensure smooth functioning of the stochastic gradient algorithms running in the background~\cite{kuhn}. It is therefore important to understand how the batch decomposition affects the overall performance. Recently in~\cite{ganesan2}, batch decompositions have been studied from a probabilistic perspective and minimum size of decompositions have been obtained for \emph{random} datasets with primarily continuous entries. In the follow up paper~\cite{ganesan3}, we obtain decomposition bounds for random categorical datasets under the constraint that similar data points are distributed across batches.

It is also important to study the effect of decompositions in \emph{deterministic} datasets, particularly in this age of large language models (LLMs) that use extensively texts broken down into small chunks for training~\cite{liu}~\cite{kazem}. In such cases, the actual distribution of datasets may be had to determine and it is of interest to establish ``distribution-free" decomposition bounds. Decompositions of deterministic datasets have been studied before in~\cite{zupan} from the context of housing loan allocations and the goal there is to improve comprehensibility and thereby study the effect on performance.

In this paper, we study decompositions of \emph{deterministic} datasets with the aim of \emph{reducing similarity} across batches.  We consider pairwise dissimilar decompositions with constraints on the number of similar pairs of data points between two batches and use random graphs to obtain ensemble bounds for the minimum size, that guarantees the existence of a reduced similarity decomposition. We then consider a relaxation of unique batch decomposition and again use Binomial random graphs to estimate the maximum size of a decomposition with a given relaxation parameter. The advantage of our approach here is that the bounds do not depend on the actual distribution of datasets which is very hard to obtain in practice.

The paper is organized as follows: In Section~\ref{sec_data}, we study dataset decompositions that aim to reduce redundancy \emph{across} batches and obtain ensemble bounds for the minimum size, that hold for nearly all redundancy graphs containing a given number~\(m\) of edges. Next in Section~\ref{sec_ubd}, we obtain ensemble bounds for the unique batch decomposition of deterministic datasets. In both cases, we obtain decomposition bounds in terms of the maximum number of data points that are similar to a given data point.

\renewcommand{\theequation}{\arabic{section}.\arabic{equation}}
\setcounter{equation}{0}
\section{Pairwise Dissimilar Decompositions}\label{sec_data}
For integers~\(L \geq 1\) and~\(1 \leq l \leq L\) let~\({\cal W}_l\) be any finite set and for~\(1 \leq i \leq n\) we let~\(W_i = (W_i(1),\ldots,W_i(L))\) be~\(n\) vectors such that~\(W_i(l) \in {\cal W}_l\) for each~\(i.\) We refer to~\(W_i\) as the~\(i^{th}\) data point or the data point at \emph{index}~\(i\) and define~\(\{W_i\}_{1 \leq i \leq n}\) to be the \emph{dataset}.

The \emph{Hamming distance} between~\(W_i\) and~\(W_j\) is defined as
\begin{equation}\label{ham_dist_def_one}
d_H(W_i,W_j) := \sum_{l=1}^{L} \ind(W_i(l) \neq W_j(l)),
\end{equation}
where~\(\ind(.)\) refers to the indicator function and denotes the number of entries where~\(W_i\) and~\(W_j\) differ. Having defined a notion of distance between data points, we now define similarity between data points as follows. For~\(0 < \beta  <1,\) we say that data points~\(W_i\) and~\(W_j\) are~\(\beta-\)similar to each other  (or simply similar) if
\begin{equation}\label{red_cond}
d_H(W_i,W_j) \leq L(1-\beta);
\end{equation}
i.e., if~\(W_i\) and~\(W_j\) agree in at least~\(L\beta\) entries.

A batch decomposition of the dataset~\(\{W_v\}\) is a partition of the indices~\(\{1,2,\ldots,n\}\) into mutually disjoint sets~\(\{{\cal V}_l\}_{1 \leq l \leq t}\) and we define~\(t\) to be the number of batches. We say that data point~\(W_v\) belongs to the~\(k^{th}\) batch if~\(v \in {\cal V}_k.\)
\begin{definition}\label{def_one}
For integer~\(r \geq 2\) we say that~\(\{{\cal V}_k\}\) is an~\(r-\)\emph{good decomposition} if the following holds for each~\(1 \leq k \leq t\) and each index~\(v \in {\cal V}_k:\) In any batch~\({\cal V}_j, j \neq k,\) there are at most~\(r\) indices whose corresponding data points are similar to~\(W_v.\)

We define~\(\Lambda_{good}(r)\) be the minimum number of batches in an~\(r-\)good decomposition of the dataset~\(\{W_v\}.\)
\end{definition}
In other words,~\(r-\)good batch decompositions ensure that any set of similar data points that recur across batches has size at most~\(r.\)



In our main result below, we estimate~\(\Lambda_{good}(r)\) in terms of the number of edges of the \emph{redundancy graph}~\(\Gamma = \Gamma(\beta)\) of the dataset~\(\{W_v\}\) obtained as follows. The graph~\(\Gamma\) is a subset of the complete graph~\(K_n\) with vertex set~\(\{1,2,\ldots,n\}\) and an edge~\((u,v)\) with endvertices~\(u\) and~\(v\) is present in~\(\Gamma\) if and only if~\(W_u\) and~\(W_v\) are similar.

For integer~\(m \geq 1\)  let~\({\cal G}_{n,m}\) be the set of all subgraphs of~\(K_n\) containing~\(m\) edges and define~\(\mathbb{P}_{n,m}\) to be the uniform distribution on~\({\cal G}_{n,m}.\) Constants throughout do not depend on~\(n\) and for sequences~\(\{a_n\}\) and~\(\{b_n\},\) we use the notation~\(a_n = o(b_n)\) to denote that~\(\frac{a_n}{b_n} \longrightarrow 0\) as~\(n \rightarrow \infty.\)
\begin{theorem}\label{thm_good}  Suppose the redundancy graph~\(\Gamma\) contains~\(m\) edges. There is a constant~\(C = C(r)  > 0\) such that
\begin{equation}\label{glass}
\mathbb{P}_{n,m}\left(\Lambda_{good}(r) \leq C\lambda_{m,n}^{\frac{r+2}{r-1}}\right)  = 1-o(1),
\end{equation}
where~\(\lambda_{m,n} := \max\left(\frac{m}{n},\log{n}\right).\)
\end{theorem}
In particular if~\(m = o\left(n^{2-\frac{3}{r+2}}\right),\) then we get that~\(\Lambda_{good}(r) = o(n)\) with high probability, i.e., with probability converging to one as~\(n\rightarrow \infty\) and so the resulting decomposition is non-trivial with high probability as well.

In our proof of Theorem~\ref{thm_good} below, we  use the following version of the Lov\'asz Local Lemma.
\begin{lemma}\label{lemmax}
Let~\(A_1,\ldots,A_t\) be events in an arbitrary probability space and suppose~\(\mathbb{P}(A_i) \leq q_A\) for all~\(i.\) Also  suppose that each event~\(A_i\) is mutually independent of all but at most~\(d_A\) of the events in~\(\{A_j\}_{j \neq i}.\) If
\begin{equation}\label{galmar}
eq_A(d_A+1) \leq 1
\end{equation}
then~\(\mathbb{P}\left(\bigcap_{i=1}^{n} A^c_i\right) > 0\) strictly.
\end{lemma}
For a proof of Lemma~\ref{lemmax},  we refer to  Corollary~\(5.1.2,\) pp.~\(65\) of~\cite{alon}.

We also use the following standard deviation estimate regarding sums of Bernoulli random variables:  Let~\(\{U_j\}_{1 \leq j \leq r}\) be independent Bernoulli random variables with~\(\mathbb{P}(U_j = 1) = 1-\mathbb{P}(U_j = 0) > 0.\) If~\(T_r := \sum_{j=1}^{r} U_j,\theta_r := \mathbb{E}T_r\) and~\(0 < \gamma \leq \frac{1}{2},\) then
\begin{equation}\label{conc_est_f}
\mathbb{P}\left(\left|T_r - \theta_r\right| \geq \theta_r \gamma \right) \leq 2\exp\left(-\frac{\gamma^2}{4}\theta_r\right)
\end{equation}
for all \(r \geq 1.\) For a proof of~(\ref{conc_est_f}), we refer to Corollary A.1.14, pp. 312 of~\cite{alon}.

\emph{Proof of Theorem~\ref{thm_good}}: We prove a version of~(\ref{glass}) for Bernoulli random graphs and then translate our bounds to the case of uniformly random graphs. Let~\(\{Z(f)\}_{f \in K_n}\) be independent random variables indexed by the edge set of~\(K_n\) and with distribution
\begin{equation}\label{x_dist}
\mathbb{P}(Z(f) = 1) = p = 1-\mathbb{P}(Z(f) = 0)
\end{equation}
for some~\(0 < p  < 1.\) Let~\(G\) be the random graph formed by the union of all edges~\(f\) satisfying~\(Z(f) = 1\) and say that vertices~\(u\) and~\(v\) are similar if the edge~\((u,v)\) with endvertices~\(u\) and~\(v\) is present in~\(G.\) Letting~\(\Lambda_{good}(r,G)\) be the minimum size of an~\(r-\)good decomposition of the vertices of~\(G,\)  we show below that for every~\(\gamma > 0\) and~\(r \geq 1,\) there is a constant~\(D > 0\) such that
\begin{equation}\label{glass_two}
\mathbb{P}\left(\Lambda_{good}(r,G) \leq D\max\left(np, \log{n}\right)^{1+\frac{3}{r-1}}\right) \geq 1-\frac{1}{n^{1+\gamma}}.
\end{equation}

Indeed, suppose~\(p \geq \frac{M\log{n}}{n}\) for some constant~\(M \geq 2\) to be determined later. We begin by estimating the degrees of vertices in~\(G.\) The expected degree of a vertex~\(v\) in~\(G\) is~\((n-1)p\) and so if~\(d(v)\) is the degree of vertex~\(v\) in~\(G,\) then from the deviation estimate~(\ref{conc_est_f}) we get for~\(0 < \epsilon < \frac{1}{2}\) that
\begin{equation} \nonumber
\mathbb{P}\left(np(1-\epsilon) \leq d(v) \leq np(1+\epsilon)\right) \geq 1-e^{-D np}
\end{equation}
for some constant~\(D > 0,\) not depending on the choice of~\(M.\) Given~\(\gamma > 0\) we choose~\(M > 0\) sufficiently large to get
\begin{equation}\label{deg_estum}
\mathbb{P}\left(np(1-\epsilon) \leq d(v) \leq np(1+\epsilon)\right) \geq 1-e^{-DM\log{n}} \geq 1- \frac{1}{n^{1+\gamma}}.
\end{equation}
Fixing such an~\(M\) and letting~\[E_{deg} := \bigcap_{v} \left\{np(1-\epsilon) \leq d(v) \leq np(1+\epsilon)\right\},\] we then get from the union bound that
\begin{equation}\label{deg_est}
\mathbb{P}(E_{deg}) \geq 1-\frac{1}{n^{1+\gamma}}.
\end{equation}

We now consider the case~\(np \leq M\log{n}.\) As before let~\(d(v)\) denote the degree of vertex~\(v\) in~\(G.\) In this case~\(d(v)\) is stochastically dominated from above by a Binomial random variable with parameters~\(n-1\) and~\(\frac{M\log{n}}{n}\) and so arguing as in~(\ref{deg_estum}), we get
\begin{equation}\label{conway}
\mathbb{P}\left(d(v) \leq 2M\log{n}\right) \geq 1-e^{-DM\log{n}}
\end{equation}
for some constant~\(D >0\) not depending on the choice of~\(M.\) Given~\(\gamma > 0,\) we choose~\(M > 0\) large enough so that the final term in~(\ref{conway}) is at least~\(1-\frac{1}{n^{2+\gamma}}.\) Setting~\(E_{deg} := \bigcap_{v=1}^{n}\left\{d(v) \leq 2M\log{n}\right\},\) we then get from the union bound that
\begin{equation}\label{e_up_est}
\mathbb{P}(E_{deg}) \geq 1-\frac{1}{n^{1+\gamma}}.
\end{equation}
We henceforth assume that~\(E_{deg}\) occurs so that each vertex contains at most
\begin{equation}\label{t_def}
T := 2\max(np,M\log{n})
\end{equation}
neighbours.

For integer~\(q = q(n)\) to be determined later, let\(X_1,\ldots,X_n\) be independent and identically distributed (i.i.d.) uniform in~\(\{1,2,\ldots,q\}\) that are also independent of~\(G\) and let~\(\mathbb{P}_X\) denote the distribution of~\((X_1,\ldots,X_n).\) Assign vertex~\(i\) to batch~\(X_i\) and let~\(\{{\cal V}_k\}_{1 \leq k \leq t}\) be the resulting batch decomposition. For a vertex~\(v\) let~\(N(v)\) be the set of all vertices adjacent to~\(v\) in~\(G\) (so that the degree~\(d(v) = \#N(v)\) is simply the size of the set~\(N(v)\)) and let~\(A_v\) be the event that there are~\(r\) vertices~\(u_1,\ldots,u_r \in N(v)\) that are assigned to the same batch~\({\cal V}_l\) for some~\(1 \leq l \leq t.\)

To compute~\(\mathbb{P}_X(A_v),\) let~\(w_1,\ldots,w_r\) be \emph{any}~\(r\) vertices in~\(N(v).\) The probability that~\(X_{w_1} = X_{w_2} = \ldots = X_{w_r}\) is equal to~\(\frac{1}{q^{r-1}}\) and since~\(E_{deg}\) occurs, there are at most~\({T \choose r} \leq T^{r}\) possible choices for~\(\{w_1,\ldots,w_r\}.\) Therefore by the union bound we get
\begin{equation}\label{q_a_def}
\mathbb{P}_X(A_v) \leq \frac{T^{r}}{q^{r-1}} =: q_A.
\end{equation}

To estimate the value of~\(d_A\) in Lemma~\ref{lemmax}, we argue as follows. For distinct vertices~\(v\) and~\(u,\) the events~\(A_u\) and~\(A_v\) are dependent on each other only if the vertices~\(u\) and~\(v\) share a common neighbour in~\(G.\) Since the maximum degree of a vertex is at most~\(T,\) the number of vertices~\(u\) such that~\(A_u\) depends on~\(A_v\) is at most~\(T^2\) and so~\(d_A \leq T^2.\) Combining this with~(\ref{q_a_def}), we get that
\[eq_A(d_A+1) \leq e \cdot \frac{T^r}{q^{r-1}} \cdot (T^2+1) \leq 2e \cdot \frac{T^{r+2}}{q^{r-1}} \leq 1\] provided~\(q = \theta \cdot T^{\frac{r+2}{r-1}}\) for some large enough constant~\(\theta > 0.\)

The condition~(\ref{galmar}) in Lemma~\ref{lemmax} is satisfied and so with positive~\(\mathbb{P}_X-\)probability, there is an~\(r-\)good decomposition of the vertices of~\(G\) containing at most~\(q\) batches. The estimates~(\ref{deg_est}) and~(\ref{e_up_est}) then imply the desired bound~(\ref{glass_two}).

We now translate~(\ref{glass_two}) to the case of uniformly random graphs as follows. Let~\(H_1\) be a Binomial random graph on~\(n\) vertices with probability distribution~\(\mathbb{P}\) and having edge probability~\(p = \frac{m}{N},\) where~\(N = {n \choose 2}.\)
For any deterministic graph~\(A \in {\cal G}_{n,m}\) we have that
\begin{equation}\label{ph1}
\mathbb{P}(H_1 = A) = p^{m}(1-p)^{N-m} = \frac{m^{m}(N-m)^{N-m}}{N^N},
\end{equation}
where we recall that~\({\cal G}_{n,m}\) is the set of all graphs with~\(n\) vertices and~\(m\) edges.

Similarly if~\(H_2\) is a uniformly randomly chosen graph from~\({\cal G}_{n,m}\) with probability distribution~\(\mathbb{P}_{n,m},\) then
\begin{equation}\label{ph2}
\mathbb{P}_{n,m}(H_2 = A) = \frac{1}{{N \choose m}} = \frac{m!(N-m)!}{N!}
\end{equation}
and so again using the Stirling approximation and~(\ref{ph1}), we then have
\begin{align}
\mathbb{P}_{n,m}(H_2 = A) &\leq 4m \cdot 4(N-m) \frac{m^{m}(N-m)^{N-m}}{N^{N}}  \nonumber\\
&= 16m(N-m) \cdot \mathbb{P}(H_1 = A) \nonumber\\
&\leq 16 n^{4} \cdot \mathbb{P}(H_1 = A), \label{translate}
\end{align}
since~\(m \leq N \leq n^2.\)

Using~(\ref{translate}) and recalling that~\(p = \frac{m}{{n\choose 2}} = \frac{2m}{n(n-1)}\) we get from~(\ref{glass_two}) that
\begin{align}
&\mathbb{P}_{n,m}\left(\Lambda_{good}(r,\Gamma) \geq D \left(\frac{2m}{n-1}, \log{n}\right)^{1+\frac{3}{r-1}}\right) \nonumber\\
&\;\;\;\leq\;\;\;\;n^{4} \mathbb{P}\left(\Lambda_{good}(r,G) \geq D \left(np, \log{n}\right)^{1+\frac{3}{r-1}}\right)   \nonumber\\
&\;\;\;\leq\;\;\;\;n^{4} \cdot \frac{1}{n^{1+\gamma}}. \nonumber
\end{align}
Since~\(\gamma > 0\) is arbitrary, this obtains~(\ref{glass}) and therefore completes the proof of the Theorem.~\(\qed\)

\renewcommand{\theequation}{\arabic{section}.\arabic{equation}}
\setcounter{equation}{0}
\section{Unique Batch Decompositions}\label{sec_ubd}
In this Section, we study batch decompositions of datasets that aim to simultaneously reduce similarity within and across clusters.  Formally, let~\(\{{\cal V}_j\}_{1 \leq j \leq t}\) be a batch decomposition of size~\(t\) and  set~\(\theta(v) := j\) if~\(v \in {\cal V}_j\) i.e., if data point~\(W_v\) is assigned to the~\(j^{th}\) batch. Also let~\(\Gamma\) be the redundancy graph obtained as in the paragraph containing~(\ref{red_cond}) and let~\(N(v)\)  be the set of all neighbours of the vertex~\(v\) in~\(\Gamma.\)

We define~\[\Theta(v) := \bigcup_{w \in N(v) \cup \{v\}} \{\theta(w)\}\] to be the set of all batches to which the neighbours of~\(v,\) including~\(v\) are assigned to. We say that the batch~\(x \in \Theta(v)\) is \emph{unique} with respect to~\(v\) if there is exactly one vertex~\(w \in N_{mod}(v) \cup \{v\}\) such that~\(\theta(w) = x.\)
\begin{definition}\label{def_uni} For~\(0 < \eta \leq 1,\) we say that~\(\{{\cal V}_i\}_{1 \leq i \leq t}\) is an~\((\eta, K)-\)unique batch decomposition (UBD)  of the dataset~\(\{W_i\}_{1 \leq i \leq n}\) if for each~\(v\) containing at least~\(K\) neighbours in~\(\Gamma,\) the set~\(\Theta(v)\) contains at least~\(\eta \#N(v)\) unique batches with respect to~\(v,\) where~\(\#A\) denotes the size of~\(A.\)
\end{definition}
In other words, we seek batch decompositions such that for any data point~\(W_v\) similar to at least~\(K\) other data points, at least a fraction~\(\eta\) of the data points similar to~\(W_v,\) are assigned to \emph{unique} batches. Our definition of unique batch decomposition is inspired from the concept of injective colouring in graphs introduced by~\cite{hahn}, where each vertex belonging to a neighbourhood is given a unique colour. We introduce and interpret the additional parameters~\(\eta\) and~\(K\) as relaxation parameters with~\(\eta=K=1\) corresponding exactly to injective colouring.

Letting~\(\Lambda_{uni}(\eta,K) = \Lambda_{uni}(\eta,K,\Gamma)\) be the minimum size of an~\((\eta,K)-\)UBD, we have the following results. As before,~\(\mathbb{P}_{n,m}\) is the uniform distribution on the set of all graphs containing~\(n\) vertices and~\(m\) edges.
\begin{theorem}\label{thm_uni} Suppose the redundancy graph~\(\Gamma\) has~\(m\) edges and let~\(0 < \eta, \epsilon< 1.\) For all~\(C > \frac{1}{|\log{\eta}|}\)  and~\(K\) large, we have
\begin{equation}\label{garuda_up}
\mathbb{P}_{n,m}\left(\Lambda_{uni}(\eta,K) \leq 2C\max\left(\frac{m}{n}, n^{\epsilon}\right)\right) =1-o(1).
\end{equation}
Conversely for every~\(\alpha < 2\eta,\) there is a constant~\(J > 0\) such that if~\(m \geq Jn\log{n},\)  then
\begin{equation}\label{garuda_down}
\mathbb{P}_{n,m}\left(\Lambda_{uni}(\eta,K) \geq \frac{2\alpha m}{n}\right) =1-o(1).
\end{equation}
\end{theorem}
Thus we see that for nearly all redundancy graphs containing at least of the order of~\(n^{1+\epsilon}\) edges, the minimum  size an~\((\eta,K)-\)UBD grows  linearly as~\(\frac{m}{n},\)  with a growth constant at most~\(\frac{1}{|\log{\eta}|}\) that is monotonic in the strictness parameter~\(\eta.\) 

\emph{Proof of Theorem~\ref{thm_uni}}: As before, we first prove a version of~(\ref{garuda_up}) for Binomial random graphs  and then translate our bounds to uniformly chosen random graphs to obtain~(\ref{garuda_up}). Details follow. Let~\(G\) be the homogenous random graph with edge probability~\(p\) as defined in~(\ref{x_dist}) and let~\(\Lambda_{uni}(\eta,K,G)\) be the minimum size of an~\((\eta, K)-\)UBD of~\(G.\)  We show below that for every~\(\gamma>0,0 < \eta,\epsilon < 1\) and~\(C > \frac{1}{|\log{\eta}|}\) strictly, we have
\begin{equation}\label{er_graph_est}
\mathbb{P}\left(\Lambda_{uni}\left(\eta,K,G\right) \leq C\max(np,n^{\epsilon})\right) \geq 1-\frac{1}{n^{1+\gamma}}.
\end{equation}

We consider two cases depending on whether~\(np\) is larger or smaller than order of~\(\log{n}.\)\\
\emph{\underline{Case I}}: Suppose that~\(np \geq M \log{n}\) for some constant~\(  M > 0\) to be determined later. For~\(\epsilon > 0\)  let~\(E_{deg}\) be the event defined prior to~(\ref{deg_est}) so that each vertex has at most~\(np(1+\epsilon)\) neighbours in~\(G.\) If~\(M > 0\) is large enough, then from~(\ref{deg_est}) we see that~\(E_{deg}\) occurs with high probability. We fix such an~\(M\) henceforth and assume that~\(E_{deg}\) occurs.

For a constant~\(C > 0\) to be determined later, we let~\(V_i, 1 \leq i \leq t = Cnp\) be~\(t\) disjoint vertex sets, each containing~\(\frac{1}{Cp}\) vertices, assuming for simplicity that~\(\frac{1}{Cp}\) is an integer. If~\(\frac{1}{Cp}\) is not an integer, then we assign the rest of the vertices outside~\(\{V_j\}_{1 \leq j \leq t}\) to a new set~\({\cal V}_{t+1}.\)  The probability that a vertex~\(v \notin V_1\) contains a unique neighbour in~\(V_1\) is~\[\alpha := \frac{1}{Cp} \cdot p \cdot (1-p)^{\frac{1}{Cp}-1} = \frac{1}{C}e^{-1/C}(1+o(1))\] for all~\(n\) large, since~\(p = o(1)\) i.e.,~\(p = p(n) \longrightarrow 0\) as~\(n \rightarrow \infty.\) Therefore if~\(X_{uni} = X_{uni}(v)\) is the number of sets in~\(\{V_j\}_{1 \leq j \leq t}\) containing a unique neighbour of~\(v,\) then~\(X_{uni}\) is Binomially distributed with parameters~\(t\) and~\(\alpha\) and from the deviation estimate~(\ref{conc_est_f}), we get for~\(\epsilon > 0\) that
\begin{equation}\nonumber
\mathbb{P}\left(X_{uni} \leq t \alpha(1-\epsilon)\right) \leq e^{-D_1t\alpha} \leq e^{-D_2np}
\end{equation}
for some constants~\(D_1,D_2 > 0\) since~\(t = Cnp\) and~\(\alpha \geq \frac{e^{-1/C}}{2}\) for all~\(n\) large.

Recalling that~\(np \geq M\log{n}\) we further get
\[\mathbb{P}(X_{uni} \leq t\alpha(1-\epsilon)) \leq e^{-D_2M\log{n}} \leq \frac{1}{n^{\gamma+2}}\] provided~\(M > 0\) is large enough. Fixing such an~\(M\) and defining~\[E_{good} :=\bigcup_{v} \left\{X_{uni}(v) \leq \theta t \alpha\right\},\]  we then get from the union bound that
\begin{equation}\label{e_good_est}
\mathbb{P}\left(E_{good}\right) \geq 1-\frac{1}{n^{1+\gamma}}.
\end{equation}
Finally, letting~\(E_{tot} := E_{good}\cap E_{deg}\) we get from~(\ref{deg_est}),~(\ref{e_good_est}) and the union bound that
\begin{equation}\label{e_tot_est}
\mathbb{P}\left(E_{tot}\right) \geq 1- \frac{2}{n^{1+\gamma}}.
\end{equation}

Suppose~\(E_{tot}\) occurs so that each vertex has at least~\(t \alpha(1-\epsilon) = npe^{-1/C}(1-\epsilon)\) unique neighbours in the sets~\(\{V_j\}_{1 \leq j \leq t}.\) Because~\(E_{deg}\) also occurs, we know that vertex~\(v\) has degree~\(d(v) \leq np(1+\epsilon)\) and so at least~\(\frac{e^{-1/C}(1-\epsilon)}{1+\epsilon} d(v) \geq \eta d(v)\) neighbours of~\(v\) are assigned to unique batches, provided~\(C > \frac{1}{|\log{\eta}|}\) strictly and~\(\epsilon > 0\) is chosen sufficiently small. Fixing such an~\(\epsilon\) and recalling that~\(\gamma >0\) is arbitrary, we get~(\ref{er_graph_est}) from the estimate~(\ref{e_tot_est}). This completes the proof of case~\(I.\)

\emph{\underline{Case II}}:  In this case we assume that~\(np \leq M\log{n}\) and  let~\(E_{deg}\) be the event as defined prior to~(\ref{e_up_est}) so that each vertex has at most~\(2M\log{n}\) neighbours in~\(G.\)

Assuming~\(E_{deg}\) occurs, we now assign batches to the data points randomly from the set~\(\{1,2,\ldots,q\},\) where~\(q \geq 1\) is a large integer to be determined later. Formally, we let~\(\{X_i\}_{1 \leq i \leq n}\) be independent random variables uniformly chosen from~\(\{1,2,\ldots,q\}\) that are also independent of the dataset~\(\{W_i\}\) and assign index~\(i\) to batch~\(X_i.\) We also let~\(\mathbb{P}_X\) denote the distribution of~\((X_1,\ldots,X_n)\) and denote~\(\{{\cal U}_{l}\}_{1 \leq l \leq q}\) to be the resulting batch decomposition.

For the data point~\(W_v,\) we set~\(d= d(v) \geq K\) to be the number of data points to which~\(W_v\) is similar to and estimate the~\(\mathbb{P}_X-\)probability that there at most~\(\eta d\) unique batches in the set~\(N(v).\) Indeed, if~\(k \leq \eta d\) indices in~\(N(v)\) are assigned to unique batches, then removing these indices from~\(N(v),\) we are left with~\(d-k\) indices assigned a total of at most~\(\frac{d-k}{2}\) distinct batches.

The number of ways of choosing the~\(k\) distinct batches from amongst a total of~\(q\) possibilities is~\({q \choose k} \) and the number of ways of assigning  indices in~\(N(v)\) to these batches is at most~\(d^{k}.\) Similarly, the number of ways of choosing~\(l \leq \frac{d-k}{2}\) batches from the remaining~\(q-k\) possibilities is~\({q-k \choose l} \leq  {q \choose l} \leq {q \choose \frac{d-k}{2}},\) by the monotonicity of the Binomial coefficient and the fact~\(d= d(v) \leq \Delta \leq \frac{q}{2}.\) The number of ways of assigning these~\(l \leq \frac{d-k}{2}\) batches to the remaining~\(d-k\) indices (allowing for multiplicity) is at most~\(\left(\frac{d-k}{2}\right)^{d-k}.\)

Since each batch assignment to the indices of~\(N(v)\) occurs with probability~\(\frac{1}{q^{d}}\) and~\(l \leq d,\) we see that the~\(\mathbb{P}_X-\)probability that there at most~\(\eta d\) unique batches in~\(N(v)\) is at most~\[\zeta := \frac{d}{q^{d}} \cdot \sum_{k \leq \eta d} {q \choose k} d^{k} \cdot {q \choose \frac{d-k}{2}} \left(\frac{d-k}{2}\right)^{d-k}.\] Using~\({q \choose l} \leq \left(\frac{qe}{l}\right)^{l},\) we get that the term
\begin{align}
{q \choose \frac{d-k}{2}} \left(\frac{d-k}{2}\right)^{d-k} &\leq \left(\frac{2eq}{d-k}\right)^{\frac{d-k}{2}} \left(\frac{d-k}{2}\right)^{d-k} \nonumber\\
&= \left(\frac{qe(d-k)}{2}\right)^{\frac{d-k}{2}} \nonumber\\
&\leq (qed)^{\frac{d-k}{2}} \nonumber
\end{align}
and so
\begin{equation}
\zeta \leq d \sum_{k \leq \eta d} \frac{1}{q^{k}} {q \choose k}d^{k} \cdot  \left(\frac{ed}{q}\right)^{\frac{d-k}{2}} \leq  d \left(\frac{ed}{q}\right)^{\frac{d-\eta d}{2}} \sum_{k \leq r} \frac{1}{q^{k}} {q \choose k}d^{k} \nonumber
\end{equation}
since~\(k \leq \eta d.\) Further using~\({q \choose k} \leq \frac{q^{k}}{k!},\) we get that
\begin{align}
\zeta &\leq d \left(\frac{ed}{q}\right)^{\frac{d-\eta d}{2}}  \sum_{k} \frac{d^{k}}{k!} \nonumber\\
&\leq de^{d} \left(\frac{ed}{q}\right)^{\frac{d-\eta d}{2}} \nonumber\\
&= d \left(e \left(\frac{ed}{q}\right)^{\frac{1-\eta}{2}}\right)^{d}.\label{stich}
\end{align}

Recalling that~\(E_{deg}\) occurs, we know that~\(d = d(v) \leq 2M\log{n}\) where~\(M  >0\) is a constant as determined prior to~(\ref{e_up_est}) and by choice~\(d \geq K.\) Given~\(0 < \epsilon < 1,\) we therefore set~\(q = Cn^{\epsilon}\) where~\(C > \frac{1}{|\log{\eta}|}\) is as in Theorem statement and get from~(\ref{stich})  that
\begin{equation}
\zeta \leq 2M\log{n} \left(e\left(\frac{2eM\log{n}}{Cn^{\epsilon}}\right)^{(1-\eta)/2}\right)^{K} \leq \frac{1}{n^{2}},
\end{equation}
provided~\(K > 0\) is large enough. Fixing such a~\(K,\) we see that with~\(\mathbb{P}_X-\)probability at least~\(1-\frac{1}{n^2},\) at least~\(\eta d\) data points with indices in~\(N(v)\) are assigned to unique batches. Consequently, if~\(F_{uni}\) is the event that there are at least~\(\eta d(v)\) unique batches in each~\(N(v),\) then an application of the union bound gives us~\[\mathbb{P}_X(F_{uni}) \geq 1 -\frac{n}{n^2} \longrightarrow 1\] as~\(n\rightarrow \infty.\)  Thus~\(F_{uni}\) occurs with high~\(\mathbb{P}_X-\)probability and this guarantees the desired~\((\eta,K)-\)unique batch decomposition containing at most~\(q\) batches. This completes the proof of~(\ref{er_graph_est}) for case~\((II).\)

As before, we now translate the estimate~(\ref{er_graph_est}) to the case of uniformly random graphs using the estimate~(\ref{translate}) obtained in the proof of Theorem~\ref{thm_good}. Indeed, from~(\ref{translate}) and~(\ref{er_graph_est}), we get that
\[\mathbb{P}_{n,m}\left(\Lambda_{\eta}(\Gamma) \geq C\Delta\right) \leq  16n^{4} \cdot \frac{1}{n^{1+\gamma}}  = \frac{16}{n^{\gamma-3}}.\] Since~\(\gamma > 0\) is arbitrary, this completes the proof of~(\ref{garuda_up}) in Theorem statement.

To prove the lower bound in~(\ref{garuda_down}), we again assume that~\(np \geq M\log{n}\) and for~\(0 < \epsilon < \frac{1}{2},\)  consider the event~\(E_{deg}\) defined prior to~(\ref{deg_est}) so that each vertex contains at least~\(np(1-\epsilon) > K \) neighbours. This necessarily implies that~\(\Lambda_{uni}(\eta,K, G) \geq \eta np(1-\epsilon)\) and so given~\(\alpha < 2\eta,\) we set~\(p = \frac{m}{{n \choose 2}}\) and choose~\(\epsilon > 0\) sufficiently small to get that~\(\Lambda_{uni}(\eta,K, G) \geq 2\alpha \frac{m}{n}.\) Again using the translate~(\ref{translate}) and the estimate~(\ref{deg_est}) for~\(E_{deg}\) and arguing as above we get the lower bound~(\ref{garuda_down}). This completes the proof of the Theorem.~\(\qed\)

\emph{\underline{Future Work}}: In this paper, we have obtained theoretical bounds for the minimum size of decompositions designed to reduce similarity across batches. In future, we plan to study practical implementations of the same and identify class of datasets that perform close enough to the bounds derived in the paper.

\subsection*{Acknowledgement}
I thank Professors C. R. Subramanian, A. Ganesh and the referee for crucial comments that led to an improvement of the paper. I also thank IMSc and IISER Bhopal for my fellowships. This work was supported by the Engineering and Physical Sciences Research Council [Grant Ref: EP/Y028732/1].


\begin{thebibliography}{99}
\bibitem{alon} N. Alon and J. Spencer. (2008).
\newblock{\em The Probabilistic Method}.
\newblock{Wiley Interscience}.

\bibitem{ganesan} G. Ganesan. (2022).
\newblock{Cliques and Chromatic Number in Multiregime Random Graphs}.
\newblock{\em Sankhya A}, \textbf{84}, pp. 509--344.

\bibitem{ganesan2} G. Ganesan. (2024).
\newblock{Dissimilar Batch Decompositions of Random Datasets}.
\newblock{\em Accepted for publication in Sankhya A}.


\bibitem{ganesan3} G. Ganesan. (2025).
\newblock{Reduced Similarity Decomposition and Subsets of Random Categorical Datasets}.
\newblock{\em Accepted for publication in Sankhya A}.

\bibitem{hahn} G. Hahn, J. Kratochv\'il,  J. Sir\'an and D. Sotteau (2002).
\newblock{On the injective chromatic number of graphs}.
\newblock{\em Discrete Mathematics}, \textbf{256}, pp. 179--192.

\bibitem{huffman} W. C. Huffman and V. Pless. (2010).
\newblock{\em Fundamentals of Error-Correcting Codes}.
\newblock{Cambridge University Press}.

\bibitem{kazem} A. Kazemnejad, I. Padhi, K. N. Ramamurthy, P. Das and S. Reddy. (2024).
\newblock{The Impact of Positional Encoding on Length Generalization in Transformers}.
\newblock{\em Advances in Neural Information Processing Systems}, \textbf{36}, 2024.

\bibitem{kuhn} M. Kuhn  and K. Johnson. (2013).
\newblock{\em Applied Predictive Modeling}.
\newblock{Springer}.

\bibitem{liu} C. Li, M. Zhang, and Y. He. (2022).
\newblock{The Stability-Efficiency Dilemma: Investigating Sequence Length Warmup for Training GPT Models}.
\newblock{\em Advances in Neural Information Processing Systems}, \textbf{35}, pp. 26736--26750.

\bibitem{zupan} B. Zupan, M. Bohanec, I. Bratko and B. Cestnik. (1997).
\newblock{A Dataset Decomposition Approach to Data Mining and Machine Discovery}.
\newblock{\em Proceedings of the Third International Conference on Knowledge Discovery and Data Mining, KDD 97}, pp. 299--302.

\end{thebibliography}
\end{document}